\documentclass{sn-jnl} 

\usepackage{graphicx}%
\usepackage{multirow}%
\usepackage{amsmath,amssymb,amsfonts}%
\usepackage{amsthm}%
\usepackage{mathrsfs}%
\usepackage[title]{appendix}%
\usepackage{xcolor}%
\usepackage{textcomp}%
\usepackage{manyfoot}%
\usepackage{booktabs}%
\usepackage{algorithm}%
\usepackage{algorithmicx}%
\usepackage{algpseudocode}%
\usepackage{listings}%
\usepackage{tikz}
\usepackage{natbib} 
\usepackage{pifont}

\begin{document}
\title{A Review on Coarse to Fine-Grained Animal Action Recognition}

\author*[1,2,3]{\fnm{Ali} \sur{Zia}}\email{ali.zia@csiro.au}

\author[2]{\fnm{Renuka} \sur{Sharma}}\email{Renuka.Sharma@csiro.au}

\author[2]{\fnm{Abdelwahed} \sur{Khamis}}\email{Abdelwahed.Khamis@csiro.au}

\author[1,2]{\fnm{Xuesong} \sur{Li}}\email{Xuesong.Li@csiro.au}

\author[2,4]{\fnm{Muhammad} \sur{Husnain}}\email{Muhammad.Husnain@csiro.au}

\author[5]{\fnm{Numan} \sur{Shafi}}\email{numan.shafi@uet.edu.pk}

\author[1]{\fnm{Saeed} \sur{Anwar}}\email{Saeed.Anwar@anu.edu.au}

\author[2]{\fnm{Sabine} \sur{Schmoelzl}}\email{Sabine.Schmoelzl@csiro.au}

\author[1]{\fnm{Eric} \sur{Stone}}\email{Eric.Stone@anu.edu.au}

\author[2]{\fnm{Lars} \sur{Petersson}}\email{Lars.Petersson@csiro.au}

\author[2]{\fnm{Vivien} \sur{Rolland}}\email{Vivien.Rolland@csiro.au}

\affil*[1]{\orgdiv{College of Science and School of Computing}, \orgname{Australian National University}, \orgaddress{\city{Canberra}, \state{ACT}, \postcode{2601}, \country{Australia}}}

\affil[2]{\orgname{Commonwealth Scientific and Industrial Research Organisation (CSIRO)}, \orgaddress{\city{Canberra}, \state{ACT}, \postcode{2601}, \country{Australia}}}

\affil[3]{\orgdiv{School of Computing, Engineering and Mathematical Sciences}, \orgname{La Trobe University}, \orgaddress{\city{Melbourne}, \state{VIC}, \postcode{3086}, \country{Australia}}}

\affil[4]{\orgdiv{School of Information \& Communication Technology}, \orgname{Griffith University}, \orgaddress{\city{Brisbane}, \state{QLD}, \country{Australia}}}

\affil[5]{\orgdiv{Department of Computer Science}, \orgname{University of Engineering and Technology}, \orgaddress{\city{Lahore}, \country{Pakistan}}}












   \abstract{This review provides an in-depth exploration of the field of animal action recognition, focusing on coarse-grained (CG) and fine-grained (FG) techniques. The primary aim is to examine the current state of research in animal behaviour recognition and to elucidate the unique challenges associated with recognising subtle animal actions in outdoor environments. These challenges differ significantly from those encountered in human action recognition due to factors such as non-rigid body structures, frequent occlusions, and the lack of large-scale, annotated datasets. 
The review begins by discussing the evolution of human action recognition, a more established field, highlighting how it progressed from broad, coarse actions in controlled settings to the demand for fine-grained recognition in dynamic environments. This shift is particularly relevant for animal action recognition, where behavioural variability and environmental complexity present unique challenges that human-centric models cannot fully address. 
The review then underscores the critical differences between human and animal action recognition, with an emphasis on high intra-species variability, unstructured datasets, and the natural complexity of animal habitats. Techniques like spatio-temporal deep learning frameworks (e.g., SlowFast) are evaluated for their effectiveness in animal behaviour analysis, along with the limitations of existing datasets. By assessing the strengths and weaknesses of current methodologies and introducing a recently-published dataset, the review outlines future directions for advancing fine-grained action recognition, aiming to improve accuracy and generalisability in behaviour analysis across species.}

\maketitle

\section{Introduction}

Animal action recognition is an emerging field of study in machine learning~\citep{ broome2023going,kleanthous2022survey,nguyen2021video}.
The intricacies of animal behaviour necessitate a level of algorithmic sophistication capable of addressing the considerable variability present in such actions~\citep{alfasly2024auxiliary}. Coarse actions are those related to general movement patterns like walking, standing, etc., while fine-grained actions pertain to the subtleties and details of behaviour~\citep{han2024multi} such as ruminating and grooming. Fine-grained animal action recognition can yield significant insights in fields such as ethology~\citep{kleanthous2022survey}, veterinary science~\citep{feng2023progressive}, and wildlife conservation~\citep{schindler2024action} by detecting subtle behavioural changes that could indicate health issues, stress, or changes in social dynamics within animal groups. 

Animal action recognition presents unique challenges compared to human action recognition. For example, animal actions occur in far more diverse and often uncontrolled natural settings, leading to higher intra-species variability. In addition, animals lack structured social and communicative cues that are often present in human actions, making it more difficult to interpret their behaviours through models built for human action recognition. Moreover, animals possess non-rigid body structures, allowing them to bend, stretch, and contort in varied ways. For example, felines have highly flexible spines, and birds can manipulate wing movements in complex patterns, unlike humans, whose skeletal structures and joint movements are relatively predictable. This anatomical flexibility introduces additional challenges when modelling animal behaviour. The problem is further compounded by frequent occlusions caused by herd behaviour or interactions with natural environments, as well as the limited availability of large-scale annotated datasets. These distinctions necessitate the development of novel algorithms tailored to capture the complexities of animal movements and their interactions with the environment. The technological advances in this area, such as the integration of open set recognition, understanding temporal dynamics, and adapting to cross-domain challenges, can push the boundaries of what machine learning and computer vision can achieve. These enhancements not only improve our ability to address specific domain issues but also contribute to the development of more sophisticated and adaptable AI systems.


\begin{figure}[tbp]
\centerline{\includegraphics[width=1\linewidth]{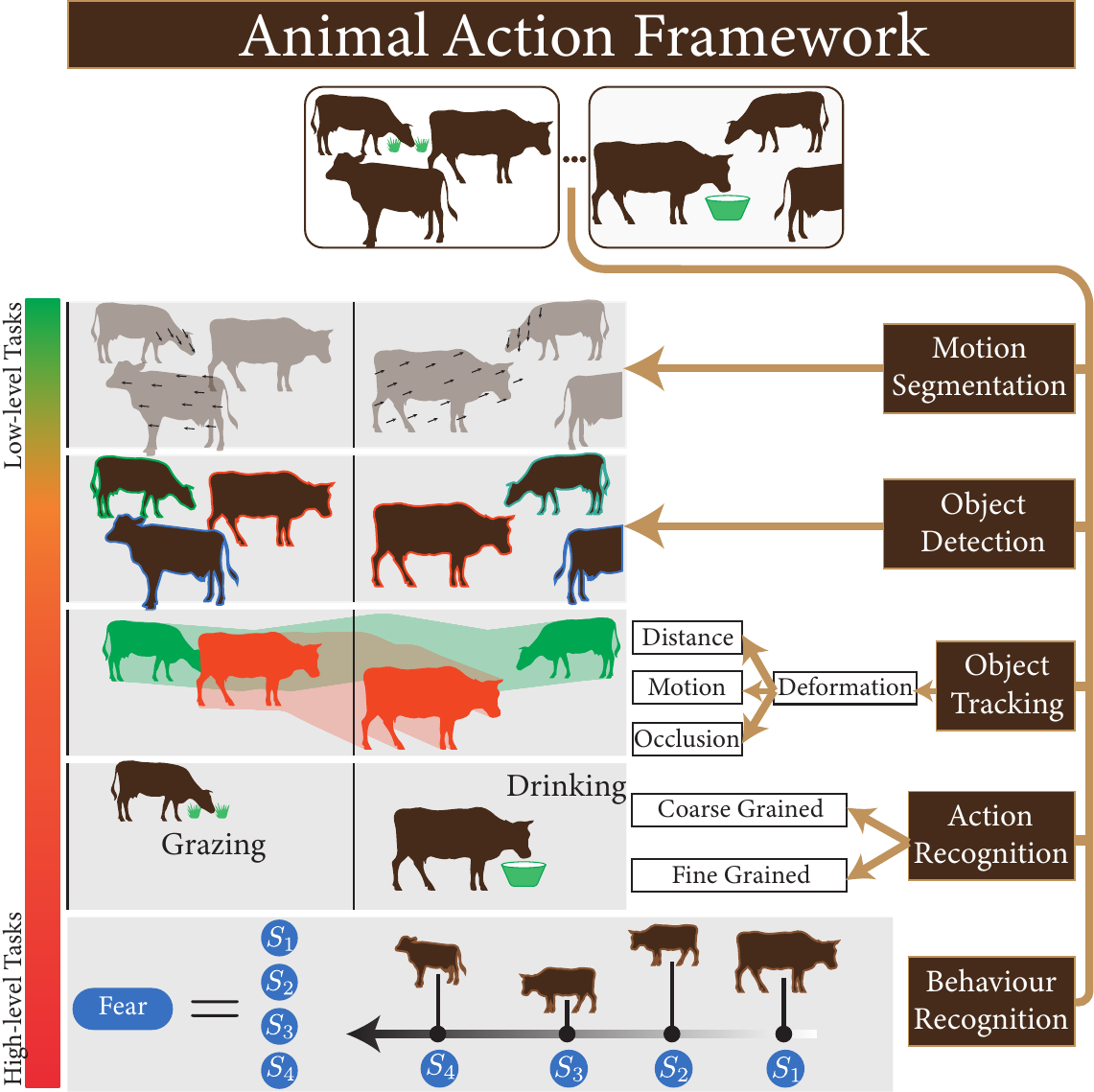}}
\caption{A general framework for animal action recognition.'S' stands for state}
\label{fig:framework}
\end{figure}

Figure~\ref{fig:framework} shows an overall framework for action recognition and behaviour understanding through a layered approach to address the granularity of actions. At the foundational level, low-level tasks such as motion segmentation and object classification are critical for initial animal detection, a process complicated by challenges such as lack of distinctive features, occlusions, varying illumination, camera angles, and types of background. These are followed by mid-level tasks, such as animal tracking, which enable higher-order analysis. Animal tracking and re-identification are complicated by deformation, with occlusions and camera-to-animal distances distorting perceived shapes. Additionally, the appearance of an animal varies with movement and viewing angles, which introduces further complexity for re-identification. High-level tasks involve recognising actions and behaviours, differentiated into coarse and fine-grained actions. This framework encapsulates a step-wise approach to decompose the progression from basic detection to complex behaviour understanding, reflecting the comprehensive methodology required to accurately interpret and analyse the spectrum of animal actions within a dynamic environment.

Fine- and coarse-grained action recognition facilitates the identification of animal behaviours by systematically analysing observed actions over extended periods through expert evaluation. \textit{Action} here, is defined as an atomic, transient event that is more specific and momentary, like grazing or running~\citep{feichtenhofer2020x3d, li2022mvitv2, tran2015learning}. These are the observable units that, when analysed collectively and over time, contribute to the understanding of more complex behaviours. \textit{Behaviour}, on the other hand, is conceptualised as the animal's response to various stimuli, observed over time, and characterised by a complex interplay of multiple actions~\citep{chen2023mammalnet, Ng_2022_CVPR}. This perspective encapsulates the holistic, ongoing nature of how animals react to their environment and internal states, often categorised under broader labels like "fear" or "stress."  The in-depth assessment of behaviours is inherently more complex than that of actions, but it can significantly benefit from advancements in action recognition. This linkage underscores that fine-grained detection and classification of discrete actions can serve as a foundational mechanism for more sophisticated behaviour monitoring. Classifying behaviour within this framework presents significant challenges. Decomposing behaviour into discrete actions or tasks is not inherently intuitive and often necessitates specialised domain knowledge or the development of dedicated models for accurate action identification. This process requires an integrated approach that combines observational acuity with analytical precision to ensure effective behaviour classification.

Although coarse action recognition has been rather well researched~\citep{Ng_2022_CVPR,joska2021acinoset}, fine-grained action recognition has only recently begun to gain attention due to its ability to capture important behavioural details~\citep{feng2023progressive,shalini2023}.
Fine-grained action recognition involves analysing minute differences in posture, movement sequences, interactions, and even facial expressions or gestures of animals to allow a deeper understanding of animal behaviour, including social interactions, emotional states, and responses to environmental stimuli. For example, while walking or standing are coarse actions, specific behaviours such as ruminating or grooming are considered fine-grained actions. 

The challenges of fine-grained animal action recognition are manifold, requiring a systematic approach to data collection, processing \citep{atto2020timed}, and analysis. Identifying individual animals and their behaviours from data is a complex task that integrates multiple aspects, demanding not only sophisticated algorithms but also robust, well-annotated datasets that reflect the variability of the natural environments in which animals are observed.
Existing datasets face challenges, including data scarcity, complex annotations, intra-class variability, occlusions, diverse environments, limited contextual information, high computational demands, and the difficulty of generalising to novel conditions~\citep{Ng_2022_CVPR}. 

To address some of these challenges, we review eight existing datasets including, a recently published Cattle Visual behaviours (CVB) dataset~\citep{zia2023cvb}, in which we have curated 502 video clips (see details in section~\ref{sec:datasets}). Each CVB video contains 450 densely labelled frames with 11 annotated perceptible behaviours, to alleviate data scarcity issues. This dataset covers real-world scenarios like occlusions and relies on natural lighting conditions. Additionally, to handle the high computational demands, we used a lightweight SlowFast~\citep{feichtenhofer2019slowfast} model, enabling the efficient identification of frequently occurring behaviours with good accuracy. 

This manuscript not only provides a comprehensive synthesis of the current state of animal action recognition, as discussed in section~\ref{sec:2dimentions}, but also addresses important challenges for available data in Section ~\ref{sec:datasets}.
Furthermore, this work sets the foundation for future advancements in Section~\ref{sec:discussion}, by highlighting the importance of integrating stable diffusion, single-point supervision, and foundational models as fundamental elements to advance our understanding of animal behaviour. In this paper, we aim to inspire and guide ongoing research in this rapidly evolving field.

\section{Facets of Fine-Grained Action Recognition}\label{sec:2dimentions}
This section investigates historical developments in coarse and fine-grained action recognition, conceptual frameworks, and computational models that have been instrumental in advancing the field.

\subsection{Evolution from Coarse to Fine-Grained Recognition}
\begin{figure*}[htbp]
\centerline{\includegraphics[width=1\linewidth]{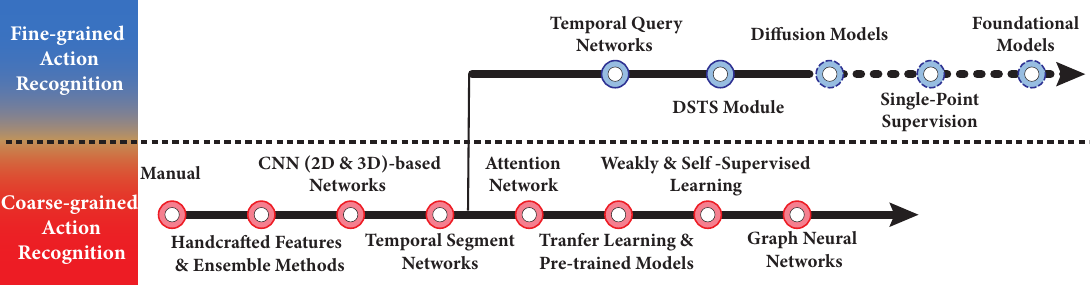}}
    \caption{
        Evolution of action recognition techniques from manual action monitoring to fine-grained action recognition.
        Above the dotted line are future opportunities to use emerging methods (section~\ref{sec:diffusion}-\ref{sec:foundational_models}).
}

\label{fig:framework-evolution}
\end{figure*}

The exploration of animal action recognition has evolved significantly, with a growing emphasis on fine-grained analysis. As shown in Figure~\ref{fig:framework-evolution}, early studies predominantly focused on coarse-grained recognition, categorising behaviours into broad groups~\citep{stern2015analyzing,Ziaeefard2015}. However, recent advances have enabled a shift towards more detailed, fine-grained recognition. This shift is well documented in the literature, with researchers such as 
\cite{lauer2022multi} and 
\cite{huang2021hierarchical} contributing significantly to the understanding of animal movements and behaviours at a more granular level. The transition to fine-grained recognition addresses the limitations of coarse-grained methods and focuses on actions at a much more detailed level, considering aspects such as individual limb movements and facial expressions.

The literature in this field reflects a diverse range of methodologies and applications but mostly caters for controlled environments. 
\cite{lauer2022multi} demonstrated the potential of DeepLabCut in multi-animal pose estimation, identification, and tracking, highlighting the advancements in pose analysis technologies. 
\cite{huang2021hierarchical} introduced a hierarchical 3D-motion learning framework, emphasising the importance of three-dimensional motion analysis in spontaneous behaviour mapping. Similarly, 
\cite{maekawa2020deep} employed deep learning-assisted comparative analysis for animal trajectories, showcasing the power of deep learning in understanding complex behavioural patterns.

The work of Stern 
\cite{stern2015analyzing} explored the use of convolutional neural networks to classify video frames, a technique that has contributed significantly to the advancement of automated behaviour recognition. 
\cite{feng2023progressive} developed a progressive deep-learning framework specifically for primate behaviour recognition, illustrating the application of these technologies in fine-grained analysis. 
\cite{kleanthous2022survey} provided a comprehensive survey of machine-learning approaches in animal behaviour, offering a broad overview of the field and its methodologies.


\cite{li2022dynamic} discuss fine-grained action recognition based on the Dynamic Spatio-Temporal Specialisation module, inspired by the human visual system. Specialised neurons learn discriminative differences for similar samples, optimised for spatial or temporal details. The Upstream-Downstream Learning algorithm enhances dynamic decisions, achieving state-of-the-art performance on two prominent datasets. Temporal Query Network \citep{zhang2021temporal} was introduced for fine-grained action understanding, leveraging a unique query-response mechanism and temporal attention. The authors employ stochastic feature bank updates for versatile training on videos of varying lengths. In the same direction, 
\cite{hacker2023fine} proposed a Two Stream Pose Convolutional Neural Network (TSPCNN) leveraging 3D CNN blocks with attention mechanisms. One stream processes raw RGB data, while the other processes Pose + RGB information. The late fusion of features yields optimal results.

\subsection{Key Considerations} 
The intricacies of fine-grained recognition necessitate taking several considerations into account, including temporal dynamics \citep{han2020tvenet} and spatial context \citep{panagiotakis2018graph}. In particular, temporal and spatial contexts are interdependent facets that, when combined, offer a comprehensive understanding of animal behaviour. 

\subsubsection{Temporal Dynamics} 
Understanding the temporal dynamics of actions involves analysing how actions evolve over time, capturing the sequence and duration of each movement. The challenge lies in detecting subtle and rapid actions, which requires advanced algorithms capable of processing high-frequency data~\citep{maekawa2020deep}. Recent studies have introduced several innovative algorithms and frameworks that are more effective in handling fine-grained actions than coarse-grained ones, particularly in capturing the nuanced progression of animal behaviour.

For instance, 
\cite{xiao2021learning} introduced a method that leverages temporal gradients as an additional modality for semi-supervised action recognition. This approach significantly improves performance by distilling fine-grained motion representations and imposing consistency across different modalities, highlighting the critical role of temporal dynamics in enhancing the accuracy of action recognition systems. 
\cite{tang2022twostream} proposed a two-stream framework that combines temporal enhanced Fisher vector encoding with graph convolutional networks for skeleton-based action recognition. This method not only preserves the temporal information of actions but also captures fine-grained spatial configurations and temporal dynamics, setting a new benchmark in the field. More recently, 
\cite{zhang2023modeling} presented Video as Stochastic Processes, a novel process-based contrastive learning framework. This framework discriminates between video processes while capturing the temporal dynamics, offering a fresh perspective on fine-grained video representation learning. Furthermore, 
\cite{Xu2024} improved animal visual tracking by introducing a spatio-temporal transformer-based method that dynamically models temporal variations to improve tracking accuracy. Their approach utilises a transformer architecture to adaptively transmit target state information across frames, effectively handling the non-rigid and unpredictable movements typical in animal behaviour. This method significantly advances the modelling and use of temporal dynamics in complex tracking scenarios.

\subsubsection{Spatial Context} 
Examining the spatial context in which an action occurs is another critical aspect of fine-grained recognition. This includes understanding the animal's interaction with its environment and other subjects or objects. Spatial context provides additional clues that can guide more accurate action recognition~\citep{huang2021hierarchical}. Recent advances in the field have introduced innovative approaches to harness spatial context effectively.

For instance, 
\cite{behera2021context} developed a context-aware attentional pooling method that captures subtle changes via sub-pixel gradients and learns to attend to informative integral regions. This approach highlights the significance of spatial context in fine-grained recognition by considering the intrinsic consistency between the information contained in the integral regions and their spatial structures. Moreover, 
\cite{hu2022sfgnet} developed a Spatial Fine-Grained Network, which leverages Spatial Fine-Grained Features by concatenating higher resolution, fine-grained features with lower resolution but semantically rich features. This method enhances object detection, particularly for small-sized objects, by incorporating spatial context through an enhanced region proposal generator and embedding contextual information surrounding regions of interest. Furthermore, 
\cite{bera2022srgnn} introduced the Spatial Relation-Aware Graph Neural Network for fine-grained image categorisation. This method aggregates context-aware features from relevant image regions and their importance in discriminating fine-grained categories, effectively utilising spatial context without requiring bounding boxes or part annotations. 

More recently, 
\cite{li2023mtfist} proposed the MT-FiST framework, a multi-task fine-grained spatio-temporal approach for surgical action triplet recognition. This framework utilises a multi-label mutual channel loss to decouple global task features into class-aligned features, thereby capturing more local details from the surgical scene. The incorporation of partial shared-parameter LSTM units to capture temporal correlations further underscores the importance of spatial context in understanding complex actions.
Furthermore, 
\cite{Xu2024} introduced an adaptive spatio-temporal inference transformer for visual tracking of coarse to fine animals. This algorithm employs a transformer-based structure to address challenges such as non-rigid deformation in animal tracking, enhancing the tracking accuracy by focussing on both coarse and fine-grained bounding box predictions. Their ablation study highlights the critical role of spatial features in improving tracking accuracy. The distribution-aware regression module alone can improve tracking metrics. When combined with coarse-to-fine tracking and target state query transmission, it significantly enhances overall performance. These results underscore the critical role of spatial features in improving tracking performance. Skeleton-based approaches have also made significant progress in capturing spatial context, particularly through multi-grained clip focus networks. These methods effectively model joint and part-level dynamics across frames, improving the spatial understanding of fine-grained actions by focussing on the spatial relationships between body parts~\citep{Qiu2024}.

\subsection{Coarse/Fine-Grained Perspectives and Integration Strategies}
Given the dynamic and varied nature of animal behaviour, it is imperative to adopt a multifaceted approach that incorporates both broad-spectrum (coarse-grained) and detailed (fine-grained) recognition strategies. Such an integrated approach is essential for developing robust and adaptable systems that can accurately reflect the intricacies of animal behaviour in various environmental contexts.

The \textit{Hierarchical framework} structures the recognition process into layers, from broad categories to specific actions, allowing for a more systematic and accurate identification process~\citep{zhao2017hierarchical}. It, therefore, implements a strategy where coarse-grained recognition (e.g. walking, running) precedes fine-grained recognition (e.g. jumping, scratching).
This approach enables the efficient processing of large datasets and optimises computational resources by involving intensive fine-grained analysis only when most relevant.

Action recognition, when viewed as an \textit{open-set problem}, involves identifying actions from known categories while also being capable of recognising previously unseen actions. This approach is crucial in dynamic environments where animals may exhibit novel behaviours not present in the training dataset~\citep{bendale2016towards}.

In \textit{action segmentation}, fine-grained recognition involves dividing continuous video streams into distinct segments, each representing a specific action. This approach is particularly challenging due to the fluid nature of animal movements and the need for precise temporal localisation of each action~\citep{lea2017temporal}. Advanced machine learning models, especially those employing temporal convolutional networks, have shown promise in effectively segmenting and recognising fine-grained actions in video data~\citep{farha2019ms,Yang2019}.

\textit{Feature fusion}, as the name indicates, fuses features extracted from coarse and fine-grained recognition techniques for comprehensive feature representation, capturing high-level context and fine details, resulting in a more robust and accurate recognition model and enhancing the method's discriminative power~\citep{shaikh2024multimodal}.

\textit{Two-stage architectures}, or cascade architectures, divide the recognition process into two stages. The coarse-grained recognition forms the first stage, and fine-grained recognition is used for cases where ambiguity persists. The two-stage model balances speed and accuracy and is vital where more detailed learning is required for challenging cases~\citep{hacker2023fine}.

With the \textit{Feedback mechanism}, coarse-grained recognition impacts the subsequent fine-grained recognition while iteratively refining the recognition, allowing for adaptive feature learning and improving model accuracy over time as feedback from fine-grained recognition influences the coarse-grained representations~\citep{yang2021feedback}.

\textit{Ensemble approaches} combine the coarse and fine-grained classifiers' outputs employing ensemble techniques like voting or weighted averaging. This provides a more reliable final result than using a single classifier as it benefits from both recognition levels and offers a balanced approach to decision-making~\citep{vu2023ensemble}.

\textit{Adaptive mechanisms} refer to techniques that adjust model behaviour dynamically based on the characteristics of the input data, allowing for more flexible and responsive action recognition. One such technique is adaptive model switching, which distinguishes between coarse and fine-grained features by assessing the complexity of the input and tailoring the recognition strategy accordingly~\citep{yang2023aim}. Similarly, recent advancements in dynamic kernel mechanisms allow models to further refine their focus on intricate details, enhancing the granularity of action recognition~\citep{Yenduri2022}.

\subsection{Modalities} 
Vision-based approaches form the cornerstone of fine-grained action recognition. These methods primarily utilise video data to analyse and classify animal behaviours. Techniques such as convolutional neural networks and deep learning algorithms have been extensively used for extracting and learning features from video frames~\citep{carreira2017quo}.

In addition to vision-based methods, other modalities such as audio, inertial sensors, and even physiological signals are being explored for fine-grained action recognition. Audio data, for example, can provide clues about vocalisations and environmental interactions, while inertial sensors can provide insights into the movement of individual animals~\citep{stowell2019automatic}.

The integration of auxiliary modalities, such as audio and text, with visual data has led to significant improvements in action recognition accuracy, especially in vision-specific datasets~\citep{Alfasly2024}. For instance, combining audio with visual data enhances the detection and classification of animal behaviour by capturing vocalisations that correspond with specific actions or states. This approach was effectively demonstrated by 
\cite{bain2021automated}, where audiovisual data was utilised to automate behaviour recognition in wild primates. Their work underscores the value of combining multimodal data, providing a more comprehensive understanding of animal behaviour and improving recognition accuracy.

Inertial sensors, often placed on animals, track their movement patterns with high precision \citep{Mao2023,Marin2020}. The data from these sensors can be used to infer detailed locomotion and activity patterns that are not easily discernible from video alone. For instance, accelerometers and gyroscopes can provide continuous data on the acceleration and orientation of animals, which, when combined with visual data, can improve the accuracy of behaviour recognition systems.

Physiological signals, such as heart rate, body temperature, and muscle activity, can also play a crucial role in fine-grained action recognition \citep{Kret2022,Broom2022,Moraes2021}. These signals can provide insights into the internal states of animals, such as stress or arousal, which are often linked to specific behaviours. The integration of such physiological data with vision-based methods can lead to a more nuanced understanding of animal actions and their underlying motivations.

Although, vision-based methods remain central to fine-grained action recognition, the incorporation of audio, inertial sensors, and physiological signals offers significant potential for enhancing the accuracy and depth of behavioural analysis. These multimodal approaches can address the limitations of single-modality systems and provide richer, more detailed insights into animal behaviour.

\section{Datasets}\label{sec:datasets}

In fine-grained action recognition, single-object datasets primarily grapple with issues like occlusions, viewpoint variations, and intra-class variability, which complicate the recognition process. This variability is exacerbated by factors such as diverse camera viewpoints, which introduce ambiguity in the recognition process. Such datasets may also suffer from limited contextual information and potential biases towards specific collection settings, challenging the generalisability of models. Conversely, multi-object datasets introduce complexities related to inter-object relationships and the inherent increase in scene complexity, which escalates computational demands and complicates tasks such as background segmentation and object tracking. The scalability of models becomes a significant concern as the number of objects increases, along with the heightened effort required for accurate labelling and annotation.
The scarcity of datasets suitable for fine-grained action recognition is a primary concern, as the meticulous capture and annotation of subtle animal behaviours demand extensive resources and expert knowledge, making data acquisition a complex task.

\begin{table}[htbp]
    \centering
    \caption{Comparison of various visual animal {\em \bf action recognition} datasets.
    }
    \vspace{1em}
    \label{tab:datasets_comparison}
    \renewcommand{\arraystretch}{1.0} 
     \small
    \begin{tabular}
{|p{0.29\textwidth}|p{0.12\textwidth}|p{0.12\textwidth}|p{0.09\textwidth}|p{0.09\textwidth}|p{0.09\textwidth}|} 
    \hline 
    \textbf{Dataset Name} & {\textbf{Multi-Object}} & \textbf{Modality}  & \textbf{CG} & \textbf{FG} & \textbf{Public}\tabularnewline
    \hline
    Large Animals \citep{liang2018benchmark}
    & \ding{55} &  Video  & \ding{51} & \ding{55} & \ding{55}\tabularnewline    
    \hline    
    Wild Felines \protect\citep{feng2021action}
    & \ding{55} &  Video  & \ding{51} & \ding{55} & \ding{55}\tabularnewline
    \hline
    Wildlife Action \citep{li2020wildlife}
    & \ding{55} &  Video  & \ding{51} & \ding{55} & \ding{55}\tabularnewline
    \hline
    Wildlife Monitoring \citep{schindler2024action}
    & \ding{55} &  Video  & \ding{51} & \ding{55} & \ding{55}\tabularnewline
    \hline
    PBRD \citep{feng2023progressive}
    & \ding{55} &  Image  & \ding{51} & \ding{51} & \ding{51}\tabularnewline
    \hline
    Animal Kingdom \citep{Ng_2022_CVPR}
    & \ding{55} &  Video  & \ding{51} & \ding{51} & \ding{51}\tabularnewline    
    \hline
    MammalNet \citep{chen2023mammalnet}
    & \ding{51} &  Video  & \ding{51} & \ding{51} & \ding{51}\tabularnewline
    \hline
    CVB \citep{zia2023cvb}
    & \ding{51} &  Video  & \ding{51} & \ding{51} & \ding{51}\tabularnewline
    \hline
    \end{tabular}  
    \renewcommand{\arraystretch}{1} 
\end{table}

In Table~\ref{tab:datasets_comparison}, we focus on eight {\em visual animal {\bf action recognition}} datasets, which are particularly relevant for the study of fine-grained behaviour.  
The table compares datasets based on the five key factors: (i)~{\em Multi-object scene} refers to the presence/absence of inter-object interactions in the scene; (ii)~{\em Modality} depicts the kind of vision modality present in the dataset; with (iii)~{\em Coarse Actions} and (iv)~{\em Fine-grained  Actions} referring to the presence/absence of such actions in each dataset; and (v)~{\em Public} if the dataset is publicly available or not.

The large animal data set~\citep{liang2018benchmark} comprises 60 cattle video recordings but does not include multi-object scenes. It shows the common actions of the cows, with manually annotated target regions for individual cows in every frame. Six prominent tracking algorithms were evaluated on this dataset to determine cow trajectories, which are essential for recognising specific actions.

The Wild Felines Dataset~\citep{feng2021action} (a collection of surveillance videos to monitor feline behaviours), the Wildlife Action Recognition dataset~\citep{li2020wildlife} (a set of animal action videos with 106 action categories), and the Wildlife Monitoring~\citep{schindler2024action} (based on camera trap videos of red deer, fallow deer and roe deer, recorded during both daytime and night-time) are datasets containing coarse actions only. The remaining four datasets reviewed here contain both coarse and fine-grained actions and are publicly available.

The Primate Fine-Grained behaviour Dataset (PBRD) introduced by 
\cite{feng2023progressive}, employs a deep CNN with a region-focused approach for identifying fine-grain behaviours across 30 classes using 7,500 image samples. This dataset significantly advances primate behaviour recognition by implementing a progressive attention training strategy that prioritises discriminative region attention. This approach progressively refines the focus on relevant image regions, allowing the model to capture and differentiate very subtle differences between similar actions. By emphasising the regions most informative for distinguishing between different behaviours, the PBRD dataset enhances the model's ability to discern subtle behavioural nuances across multiple hierarchical levels, thereby achieving superior accuracy in fine-grained action recognition.

The Animal Kingdom dataset~\citep{Ng_2022_CVPR} comprises 50 hours of annotated videos aimed at locating pertinent animal behaviour segments in lengthy videos for ground-truthing. It also includes 30,000 video sequences for fine-grained multi-label action recognition and 33,000 frames dedicated to pose estimation. This dataset encompasses a diverse array of animals, featuring 850 species spanning six major animal classes.

The MammalNet dataset~\citep{chen2023mammalnet} is built around 173 mammal categories and includes 12 common high-level mammal behaviours (e.g., hunting, grooming), making it suitable for the study of both action and behaviour recognition.

The CVB dataset~\citep{zia2023cvb}~\footnote{available at \url{https://doi.org/10.25919/3g3t-p068}.} is a real-world dataset consisting of eleven action categories with a behaviour label provided for each cattle present in a given frame. This dataset includes fine-grained behaviours such as grooming, ruminating-lying, and ruminating-standing. There are at most eight cattle (multi-object scene) in a field of $25m \times 25m$ in size in the CVB dataset. The videos were recorded from four viewpoints using a Go-Pro cameras located at each corner of the field during the day, using natural lighting conditions. This multi-view scenario introduces deformation due to distance, motion, and occlusion, representing realistic challenges of real-world datasets. Each video in the CVB dataset has 450 frames annotated by domain experts. All cattle in each frame are given a unique identifier that stays consistent throughout the video. Datasets like CVB are ideal to design novel approaches that can recognise and track cattle behaviour in real-time. Out of the eight datasets listed in Table~\ref{tab:datasets_comparison}, MammalNet and CVB are the only datasets with multi-object scenes and complexities related to inter-object relationships.

Public animal datasets are not limited to action recognition.
{\em \bf Object detection} datasets like iNaturalist~\citep{van2018inaturalist}, Animals with Attributes~\citep{xian2018zero}, Caltech-UCSD Birds~\citep{wah2011caltech}, Animals-10~\citep{narayan2022adjusting}, and Florida Wildlife Animal Trap~\citep{gagne2021florida}) focus on animal detection only.
However, some {\em \bf pose detection} datasets like 
Fish4Knowledge~\citep{spampinato2008detecting}, OpenApePoses~\citep{desai2023openapepose}, Animals with Attributes 2 (AwA2)~\citep{xian2018zero}, AcinoSet~\citep{joska2021acinoset}, Animals-10~\citep{narayan2022adjusting}, Animal Kingdom~\citep{Ng_2022_CVPR} are used for pose estimation in animals. These datasets include still images of animals, and the idea includes the localisation of the key points for behaviour identification from the particular stance of the animal in a picture, which requires domain expertise. Most of these datasets focus on a relatively small number of behaviour classes for a particular animal species/genus.

Although there are various datasets for animal action recognition, the field of fine-grained action recognition in animals is still nascent. The Animal Kingdom and PBRD datasets include valuable fine-grained actions for a few animal species but lack an exhaustive set of actions for all animal species in the dataset. Contrastingly, our CVB dataset focuses on the wide variety of actions of a single animal species, cattle (\emph{Bos taurus}), with behaviours spanning coarse actions to fine-grained actions, and as such, is a valuable contribution to the field. 

\section{Fine-Grained Behavioural Analysis: Trends, Limitations, and Emerging Techniques}\label{sec:discussion}
The shift from coarse to fine-grained recognition in animal action analysis represents a significant step towards better understanding animal behaviour. However, the scarcity of relevant datasets is not the only challenge in this space. A comprehensive examination of the existing literature uncovers limited studies focusing on fine-grained action recognition tasks. Relevant efforts in this direction are summarised in Table~\ref{tab:action-recognition}, which outlines the most recent deep learning approaches utilising the datasets referenced in Table~\ref{tab:datasets_comparison}. 
Among these, PBRD, Animal Kingdom, MammalNet, and CVB~\citep{zia2023cvb} datasets have endeavoured to recognise fine-grained actions separately on animal videos.

\begin{table*}[htpb]
\centering
\caption{Summary of different methods for wildlife and cattle action recognition. Accuracy reported as mixed indicates that the corresponding work does not report FG (Fine Grain) and CG (Corse Grain) results separately.}

\label{tab:action-recognition}
\renewcommand{\arraystretch}{1.0} 
\small
\begin{tabular}
{|p{0.08\textwidth}|p{0.27\textwidth}|p{0.25\textwidth}|p{0.28\textwidth}|}
\hline
\multicolumn{1}{|c|}{\textbf{Type}} & \multicolumn{1}{c|}{\textbf{Method}} & \multicolumn{1}{c|}{\textbf{Dataset}} & \multicolumn{1}{c|}{\textbf{Accuracy}} \\
\hline
CG & Trajectory-based Approches & Large Animals~\textbf{\citep{liang2018benchmark}} & 88.4\% - 94.1\%   \\ \hline
CG & Two-Stream Network & Wild felines dataset~\textbf{\citep{feng2021action}} & 92\% - 97\% \\ \hline
CG & I3D Hierarchical Networks & Wildlife animal dataset~\textbf{\citep{li2020wildlife}} & 33\% - 36\% \\ \hline
CG & Mask-Guided Action Recognition (MAROON) & Wildlife Monitoring\textsuperscript{\citep{schindler2024action}} & 43.05\% - 69.16\% \\ \hline
CG $+$ FG & Progressive Deep Learning Framework & PBRD~\textbf{ \citep{feng2023progressive}} & CG (48\% - 91.53\%) \newline FG (29.62\% - 81.90\%) \\ \hline
CG $+$ FG & CARe Model(I3D, X3D, and SlowFast) & Animal Kingdom~\textbf{\citep{Ng_2022_CVPR}} & Mixed (27.3\% - 39.7\%) \\ \hline
CG $+$ FG & I3D, C3D, SlowFast, and MViT V2 & MammalNet~\textbf{\citep{chen2023mammalnet}} & Mixed (34.2\% - 46.6\%) \\ \hline
CG $+$ FG & SlowFast Network & CVB~\textbf{\citep{zia2023cvb}} & CG (58\% - 73.96\%) \newline FG (12.7\% - 29.6\%) \\ \hline

\end{tabular}
\renewcommand{\arraystretch}{1} 
\end{table*}

The accuracy of the fine-grained action recognition methods on the video datasets is low due to the involvement of spatio-temporal dynamics, as evident by~\citep{zia2023cvb,Ng_2022_CVPR,chen2023mammalnet}. Progressive Deep Learning Framework \citep{feng2023progressive}, when evaluated on PBRD, demonstrated a promising potential on finer primate behaviours. However, its general applicability for fine-grained behaviours in real-world scenarios and across the animal kingdom is limited for two key reasons. Firstly, models trained on bipeds have limited transferability to quadripeds. Secondly, PBRD consists of still images with isolated fine-grained actions and does not include any fine-grained behaviour with temporal context. Therefore, the framework can not recognise context-specific fine-grained behaviours. The {\em Accuracy} column in Table~\ref{tab:action-recognition} presents the accuracies for coarse-grained and fine-grained action recognition using methods presented in the {\em Methods} column. For each dataset in the table, we list accuracy ranges based on available action granularity. The significantly lower recognition accuracies for fine-grained actions clearly indicate the complex nature of these actions.

Fine-grained action recognition-based methods use a combination of I3D~\citep{carreira2017quo}, C3D~\citep{tran2015learning}, X3D~\citep{feichtenhofer2020x3d}, MViT v2~\citep{li2022mvitv2}, and SlowFast~\citep{feichtenhofer2019slowfast} in the Animal Kingdom, MammalNet, and CVB datasets. The SlowFast method operates on two pathways: a slow pathway on a low frame rate and a fast pathway on a high frame rate, capturing spatial features and motion with fine temporal resolution, respectively. While primarily used for action recognition in CVB, it can also be applied to behaviour classification, as demonstrated in MammalNet. PBRD is a labelled fine-grained recognition dataset, and ~\citep{feng2023progressive} employs a region-focused deep convolutional neural network (Progressive Deep Learning Framework) for action classification. The model uses a progressive attention training strategy, focusing on highlighting discriminative regions and promoting complementarity across various leading levels.

While many of the reviewed models demonstrate promising results in benchmark settings, their performance often deteriorates in unconstrained, real-world environments. For example, models like I3D and C3D tend to struggle with occlusion, background clutter, and variations in lighting commonly found in field conditions. Similarly, the CARe model, although effective in classifying both coarse and fine actions in the Animal Kingdom dataset, suffers from inconsistent detection when multiple animals interact or overlap—an issue prevalent in multi-object scenarios. A significant limitation of hierarchical or cascade-based coarse-to-fine models is their dependence on accurate coarse predictions. Errors made at the coarse stage can propagate, causing the fine-grained module to misclassify actions or entirely miss subtle transitions. Moreover, many current models are trained on trimmed clips, making them ill-suited for continuous behaviour monitoring, where segmentation and recognition must occur jointly. Temporal granularity remains an unsolved challenge—models like SlowFast and MViT v2 capture short-term dynamics well but may fail to model long-term patterns or context dependencies in behaviour sequences.

Fine-grained action recognition accuracies are mediocre across the reviewed methods. This observation underscores the pressing need to develop and implement more sophisticated and precise methodologies for fine-grained action recognition. Integration of stable diffusion models, single-point supervision techniques, and large-scale foundational models is a promising avenue to address current limitations. The following subsections explain how these approaches may significantly enhance the granularity and accuracy of behavioural analyses, thereby contributing to a deeper and more nuanced understanding of animal behaviour.

\subsection{The Promise of Stable Diffusion}
\label{sec:diffusion}

The utilisation of diffusion models, particularly the stable ones, has marked a significant advance in the field of generative tasks, extending their applicability to discriminative tasks, such as object detection and image segmentation. Their core principle, iterating the denoising process to recover data samples, has been foundational in achieving stable training and generation processes.

In the context of action recognition, the application of diffusion models, as exemplified by DiffTAD~\citep{nag2023difftad} and DiffACT~\citep{liu2023diffusion}, marks a novel approach. These models employ the denoising process, conditioned on temporal location queries within videos, to accurately recover action sequences. This methodology aligns with the iterative refinement intrinsic to diffusion models, making them particularly suited for the multi-stage nature of action recognition tasks. The diffusion-based data augmentation method proposed by 
\cite{jiang2023spatial} exemplifies the innovative use of diffusion models to generate high-quality, diverse action sequences, enhancing the robustness of action recognition systems. Similarly, the action text diffusion prior network introduced by 
\cite{zhuang2023action} underscores the potential of diffusion models in improving the quality of video feature extraction, which is crucial for accurate action recognition.

The mathematical formulation of the diffusion process in action recognition involves corrupting the ground truth action $y_0$ with Gaussian noise through a series of steps, leading to $y_s = \sqrt{\hat{\alpha_s}}y_0 + \epsilon\sqrt{1-\hat{\alpha_s}}$, where $\epsilon \sim \mathcal{N}(0, I)$, $\hat{\alpha_s}$ 
calibrates the degree of Gaussian noise,
and $s$ is the random diffusion step. In the denoising step, the corrupted $y_s$ and encoded video features are given as input to a decoder $\mathcal{G}$. The decoder is then used to remove added noises and predict uncorrupted action values, i.e. $\hat{y_s} = \mathcal{G}(y_s, s, ve)$. After an iterative denoising process, the encoder $\mathcal{G}$ can predict $\hat{y_0}$.

The gradual denoising approach is instrumental in highlighting subtle distinctions crucial for fine-grained analysis, allowing the models to discern minor yet significant differences between closely related actions. The incorporation of encoded video features during denoising enriches the ability of the model to extract contextual and temporal nuances essential for understanding complex action sequences. Furthermore, the inherent adaptability of stable diffusion models, conditioned on diverse inputs, ensures their utility for fine-grained analysis. 

Despite the promise stable diffusion holds for fine-grained actions, it is not devoid of challenges. These include ensuring temporal coherence in dynamic sequences, handling high-dimensional data to capture subtle details, and dealing with complex backgrounds and occlusions that can mask critical action features. The scarcity of large, detailed datasets necessary for training these models exacerbates these challenges, along with the difficulty of generalising across the inherent variability of animal behaviours and environmental conditions. Furthermore, integrating contextual information crucial for understanding fine-grained actions, the significant computational resources required for processing high-resolution data, and achieving real-time processing amidst the iterative nature of diffusion processes present additional hurdles. Overcoming these obstacles necessitates innovative advances in model architectures, training techniques, and computational strategies to unlock the full potential of stable diffusion models for accurate and efficient fine-grained action recognition.

\subsection{Explanation of Single-Point Supervision and its Relevance}
\label{sec:siggle-point}

Single-point supervision is another emerging concept that offers promise for fine-grained recognition. It refers to the training of recognition models using minimal data points, reducing the need for extensively annotated datasets while still achieving high accuracy. This approach is particularly beneficial in scenarios where collecting large-scale, detailed annotations is impractical or impossible~\citep{bain2021automated} such as in wildlife monitoring or ethological studies. 
In their recent study, 
\cite{Yin2023ProposalBased} applied point-level supervised temporal action localisation to untrimmed videos, introducing a pioneering methodology for localising actions. This method is characterised by its generation and evaluation of action proposals with variable durations. A notable aspect of their approach is the implementation of an efficient clustering algorithm designed to generate dense pseudo-labels from point-level annotations. Additionally, the study incorporates a fine-grained contrastive loss, which plays a crucial role in enhancing the precision of these labels. In the context of fine-grained action recognition, such an approach would involve identifying specific regions within video frames, thereby providing deep learning algorithms with targeted guidance on where fine-grained actions are likely to occur. This strategy could significantly improve the accuracy and efficiency of fine-grained action recognition systems in detecting and analysing intricate animal behaviours.
Let $D = \{(x_i, y_i)\}_{i=1}^{N}$ be a dataset where $x_i$ represents the input data (such as video frames or image sequences), and $y_i$ denotes the corresponding annotations for animal actions. In traditional supervised learning, each $x_i$ requires a detailed annotation $y_i$. However, under single-point supervision, the dataset is transformed to $D' = \{(x_i, y'_i)\}_{i=1}^{N}$, where $y'_i$ are minimal, point-based annotations instead of full labels. 
The goal is to train a model $f_{\theta}$ on $D'$ so that it can predict the detailed annotations $y_i$ from these minimal data points $y'_i$. The learning objective can be expressed as optimising model parameters $\theta$ to minimise the loss function $L(f_{\theta}(x_i), y_i)$, even when trained on the sparse annotations $y'_i$.

Single-point supervision is notably advantageous in the realm of fine-grained animal action recognition for its resource efficiency. This method proves especially valuable in environments where collecting extensive annotations is impractical, such as in detailed wildlife behaviour studies. Moreover, it boasts the capability of achieving high accuracy with minimal data, a critical feature for discerning the subtle complexities inherent in animal behaviours.

Despite these strengths, the approach is not without challenges. The efficacy of single-point supervision greatly depends on the quality of initial annotations, where inaccuracies can compromise the final training outcomes. Additionally, the method may introduce increased model complexity, necessitating more sophisticated machine learning techniques and computational resources. Nonetheless, single-point supervision represents a significant opportunity towards balancing efficiency and precision in animal behaviour analysis, particularly in scenarios where data collection is difficult.

\subsection{Leveraging Foundational Models for Fine-Grained Action Recognition}
\label{sec:foundational_models}
Foundational models, such as large-scale pre-trained neural networks, have become pivotal in various domains of machine learning and computer vision due to their ability to generalise across a wide range of tasks and datasets. In the context of fine-grained action recognition, these models offer a robust framework for capturing the intricate details of animal behaviours, thanks to their deep architectural layers and extensive training on diverse data~\citep{Bilal2021TransferLB}.

Foundational models, pre-trained on vast datasets, can be fine-tuned for specific tasks like fine-grained action recognition. This transfer learning approach allows the leveraging of learned features and representations to reduce the need for large annotated datasets specific to animal actions~\citep{Bilal2021TransferLB}.
These models excel in extracting rich, hierarchical features from data, enabling the identification of subtle nuances in animal movements and behaviours that are often overlooked in coarse-grained analyses~\citep{Sun2022FineGrainedAR}.
Advanced foundational models, especially those designed for video processing, are able to capture temporal dynamics and spatial contexts, crucial for understanding the progression and environment of animal actions~\citep{Sun2022FineGrainedAR}.

In addition to transfer learning, there are other strategies to integrate foundational models with fine-grained recognition. One such approach is to combine the strengths of foundational models (for feature extraction and generalisation) with custom layers or algorithms designed for the nuances of fine-grained action recognition to optimise performance. It is also possible to use knowledge distillation mechanisms, enabling a compact student model to learn fine-grained action recognition by mimicking the output of a more complex teacher model to capture subtle action differences. 

 Test-time adaptation (TTA) is an emerging strategy to further enhance the applicability of foundational models in dynamic environments. TTA involves adapting a pre-trained model during the inference phase to better align with the statistical properties of the test data, thereby improving robustness to distribution shifts that were not seen during training. For example, ViTTA~\citep{lin2023video}, a video-tailored adaptation method, aligns the training statistics with online estimates of test statistics and enforces prediction consistency over temporally enhanced views of the same test video sample. This approach is particularly beneficial for fine-grained animal action recognition, where environmental variations and subtle differences in animal behaviour can significantly impact model performance.

Finally, another approach is to combine the capabilities of foundational models to process and analyse data from multiple modalities (e.g. visual, auditory, and sensor data) to provide a more comprehensive understanding of animals. This approach involves amalgamating features from diverse sources to enhance the accuracy and depth of recognition capabilities. For multi-modal data fusion, let $X_v, X_a, \ldots$ represent different modalities such as visual and auditory data. The fused feature set $F_{\text{fused}}$ can be obtained by a fusion function $h$:

\begin{equation}
F_{\text{fused}} = h(f(X_v; \theta_v), f(X_a; \theta_a), \ldots)
\end{equation}

Where $\theta_v, \theta_a, \ldots$ are the parameters of the feature extraction functions for each modality.

While foundational models hold great potential, their application in fine-grained action recognition comes with challenges. The complexity and size of foundational models demand significant computational power, hindering their practical applications for real-time analysis and deployment in resource-constrained environments. Additionally, models trained on general datasets may not fully capture the specificities of animal behaviours, necessitating careful fine-tuning and validation. The interpretability challenge, particularly the "black-box" nature of these models, complicates understanding their decision-making, posing challenges for validation and trust in critical applications. Despite hurdles, these models have the potential to enhance our understanding of animal behaviour significantly. Research focused on enhancing the computational efficiency and real-time applicability of these models, coupled with efforts to improve their interpretability and generalisation capabilities, holds promise for transforming wildlife monitoring, veterinary science, and ethological research.

These approaches point toward a broader paradigm shift: from narrow, task-specific classifiers to adaptive, generalist models that can leverage prior knowledge, adapt on-the-fly, and interpret subtle behavioural cues. In this sense, the future of fine-grained animal behaviour recognition aligns closely with the ongoing evolution of multimodal, foundation model-based AI systems.

\section{Challenges, Ethical Considerations, and Future Directions}
The development of fine-grained animal action recognition systems faces numerous challenges—not only in terms of technical complexity, data limitations, and generalisability but also with regard to ethical responsibility. As these systems move from controlled lab settings to real-world deployment in conservation, agriculture, and research, it is vital to reflect on the potential societal, ecological, and ethical implications. This section outlines key barriers to progress, introduces ethical considerations, and proposes future research directions to advance the field responsibly.

\subsection{Challenges}

\noindent
\textbf{The complexity of Animal Behaviours:} The inherent complexity and diversity of animal behaviours continue to pose significant challenges, particularly in terms of accurately classifying and interpreting subtle and rapid actions.

\noindent
\textbf{Data Scarcity:} While datasets such as CVB are a step forward, specialised fine-grained datasets remain scarce, limiting the scope and applicability of recognition models. This is due in part to the fact that labelling fine-grained behaviours in video data is time-consuming and laborious.

\noindent
\textbf{Technological Limitations:} The reliance on advanced technologies such as stable diffusion and single-point supervision also brings forth challenges related to computational resources, model training, and the need for specialised expertise.

\noindent
\textbf{Generalisability:} Applying findings from controlled environments to more dynamic, outdoor settings requires models that can adapt to varied and sometimes unpredictable conditions.

\noindent
\textbf{Integration of Modalities:} While significant progress has been made in vision-based recognition, integrating multiple modalities (e.g. audio, inertial sensors, etc.) for a more holistic understanding of animal behaviour is still in its infancy.

\noindent
\textbf{Real-Time Processing:} Developing systems capable of processing and analysing data in real-time to provide immediate insights into animal behaviour poses a technical challenge. This is crucial for applications requiring timely interventions, such as wildlife conservation efforts and precision livestock farming.

\noindent
\textbf{Inter-Species Variability:} Different animal species exhibit unique behaviours and physiological responses. Developing recognition models that can accurately interpret behaviours across various species remains challenging. For example, the same gesture or movement might indicate different states in different species, necessitating species-specific models or adaptable algorithms.

\noindent
\textbf{Ethological Validity:} Ensuring that the behaviour recognition systems maintain ethological validity is crucial. This involves not only identifying behaviours correctly but also understanding the context and significance of these behaviours within the species’ natural setting. Misinterpretation of behaviours due to a lack of ethological insight can lead to incorrect conclusions.

\noindent
\textbf{Data Annotation Complexity:} Fine-grained annotation of animal behaviours is labour-intensive and requires domain expertise. Unlike human activity datasets, which can often be annotated by non-experts, animal behaviour annotation often requires knowledge of subtle and species-specific actions, making the annotation process both costly and time-consuming.

\noindent
\textbf{Environmental Interference:} Natural habitats present numerous challenges, such as varying lighting conditions, occlusions from foliage or other animals, and dynamic backgrounds. These factors can significantly interfere with the accuracy of visual and sensor-based recognition systems, requiring advanced preprocessing and robust algorithms.

\subsection{Ethics and Responsible AI in Animal Action Recognition}
As the field of animal action recognition advances, it is imperative to consider the ethical implications of data collection, model deployment, and automation in sensitive ecological and agricultural settings. Wildlife monitoring, for example, often involves passive surveillance technologies that record animals without consent—raising questions about the ethical treatment of non-human subjects, particularly in protected habitats. In livestock farming, over-reliance on automated systems could reduce human oversight and potentially delay interventions when AI systems misinterpret health or behavioural cues.

Data bias is another critical concern. Many existing datasets are species-specific or collected under particular environmental conditions (e.g., daylight, fenced enclosures), resulting in models that may generalize poorly to broader ecological contexts. This bias can inadvertently reinforce human-centric interpretations of animal behaviour, overlooking subtle inter-species and intra-species differences.

Furthermore, real-time behavioural classification systems, while useful for operational efficiency, must be evaluated against potential risks such as false positives in aggression detection or incorrect stress classification. In conservation contexts, misclassifying a behavioural cue could lead to misguided interventions.

Responsible development of animal behaviour recognition systems thus requires close collaboration with ethologists, veterinarians, and ecologists. Transparent documentation of dataset provenance, model limitations, and deployment conditions should become standard practice to ensure ethical integrity and real-world impact.

\subsection{Future Directions}
Despite the advances and contributions outlined in this review, there are significant research gaps that need future exploration in the field of animal action recognition. A key limitation in the current body of research is the scarcity of fine-grained action datasets, such as the Cattle Visual Behaviours (CVB) dataset introduced in this study. The lack of comprehensive and diverse datasets across multiple species continues to hinder model generalisability and performance. To address this, future research should prioritise the development of datasets that capture fine-grained actions in a broader range of species, behaviours, and environmental contexts. For example, integrating multimodal data, such as audio and physiological measurements (e.g. heart rate, body temperature), with visual information could provide deeper insight into animal behaviours. Furthermore, existing multispecies datasets, such as MammalNet, which currently focuses on 173 mammal species, could be expanded to include a more diverse range of taxa, such as avian and aquatic species. This approach could help overcome domain adaptation challenges and improve model robustness in varied ecological settings.

Furthermore, integrating artificial intelligence with disciplines such as ethology, ecology, and veterinary science can foster a more nuanced understanding of animal behaviour. For example, AI models informed by ethological principles could help analyse how stress manifests in livestock, allowing for the early detection of diseases or changes in social dynamics. A practical application of this would be the use of artificial intelligence in precision agriculture, where models could monitor herd behaviour and detect early signs of disease in cattle based on fine-grained changes in posture, potentially transforming livestock management practices. In addition, understanding the interactions between multiple animals is crucial for studying social structures and group behaviours. In species such as elephants and primates, where social hierarchies strongly influence individual behaviour, capturing these dynamics requires advanced models capable of detecting and analysing interactions between individuals. Future research should focus on developing algorithms to analyse group interactions, such as the synchronisation of movements in herds or flocks, which could provide valuable insight into social behaviour, leadership dynamics, predator-prey interactions, and cooperative behaviours in species such as wolves or dolphins.

A further limitation of current models is their reduced accuracy in field conditions, where factors such as varying lighting, occlusions, and background complexity are common. Future research should aim to develop models capable of handling such environmental variability without significant loss in performance. For example, models trained with synthetic data could be tested in real-world environments to simulate and predict behaviours under dynamic conditions, such as those encountered in the wild or agricultural settings. Integrating environmental data, including habitat types, weather patterns, and vegetation density, could also improve behavioural analysis. For example, recognising behavioural differences between animals in arid versus temperate environments requires models that account for such contextual data. Environmental sensors that measure variables such as temperature, humidity, or vegetation indices could be incorporated into recognition systems, enabling a richer understanding of how environmental factors influence animal behaviour.

Although stable diffusion models have shown promise in iteratively refining predictions, they are computationally intensive. Future research should focus on reducing computational complexity while maintaining the precision needed for fine-grained behaviour recognition. Likewise, single-point supervision techniques, which rely on minimal labelled data, warrant further exploration to address annotation challenges in wildlife studies. For instance, 
\cite{Yin2023ProposalBased} approach to point-level temporal action localisation in untrimmed videos could be extended to detect nuanced animal behaviours over long periods and improve the efficiency of behaviour recognition systems in natural habitats. Transfer learning could also enable models trained on one species to generalise to others with minimal retraining. For example, models trained on cattle behaviour could be adapted to recognise behaviours in other ungulates, such as deer, by taking advantage of common locomotion and social patterns. Such a cross-species generalisation could significantly reduce the cost of data collection and annotation, improving the scalability of recognition systems across diverse species and environments. Moreover, developing models capable of not only recognising but also predicting future behaviours in real time remains a critical challenge. Predictive models could be instrumental in anticipating adverse events such as aggression or distress, which are crucial for both wildlife conservation and livestock management. For example, predictive models could predict health-related behaviours such as lameness or calving in cows, allowing timely interventions and minimising animal suffering.

Additionally, there is a need to extend fine-grained action recognition beyond isolated behaviour detection. Future work should focus on the longitudinal analysis of behaviour sequences, exploring how fine-grained actions contribute to more complex behavioural patterns over time. 

\section{Conclusions}
This review has presented a comprehensive analysis of the state-of-the-art techniques in animal action recognition, with a focus on the transition from coarse- to fine-grained methodologies. Our work has highlighted the importance of capturing subtle, fine-grained animal behaviours, particularly in contexts where minute changes in posture or activity may signal critical shifts in health, stress, or social dynamics. This paper has reviewed eight existing datasets in the field of animal action recognition, emphasizing that MammalNet and the CVB dataset hold particular promise for advancing research in this area. These datasets, with their integration of more naturalistic, real-world scenarios, enable more accurate modelling of animal behaviours under diverse environmental conditions. However, this review also identifies several limitations that need to be addressed. First, the generalisation of the findings across different species remains limited, as the CVB data set is focused solely on cattle. Additionally, while the SlowFast model performs well on fine-grained actions, recognition accuracies remain lower compared to coarse-grained actions, highlighting the ongoing challenge of capturing the full complexity of fine-grained behaviours.

The implications of this work extend beyond academic interest. Researchers in fields such as veterinary science, ethology, and animal production can benefit from these advances in fine-grained action recognition to better monitor and interpret animal behaviours, potentially leading to improved animal welfare and more efficient management practices. Furthermore, the introduction of novel techniques like stable diffusion, single-point supervision, and foundational models opens new avenues for the field. These technologies promise to improve accuracy while reducing the reliance on large, annotated datasets, making animal action recognition more accessible and scalable across diverse applications. Although this review has highlighted significant advancements in fine-grained action recognition, it also emphasises the need for further research, particularly in the areas of dataset expansion, real-time recognition, and cross-species generalisation. By addressing these challenges, future work will continue to push the boundaries of what is possible in animal behaviour analysis, offering new insights and practical applications across a range of scientific disciplines.

\bibliographystyle{sn-vancouver}
\bibliography{ref}
\end{document}